%% file: neurips_2024.tex
\documentclass{article}

     \usepackage[preprint]{neurips_2024}

\usepackage[utf8]{inputenc} 
\usepackage[T1]{fontenc}    
\usepackage{url}            
\usepackage{booktabs}       
\usepackage{amsfonts}       
\usepackage{nicefrac}       
\usepackage{microtype}      
\usepackage{xcolor}         
\usepackage{amsmath}
\usepackage{algorithm}
\usepackage{algpseudocode}

\DeclareMathOperator*{\argmin}{arg\,min}
\usepackage{graphicx}
\usepackage{tcolorbox}
\tcbuselibrary{most}
\usepackage{enumitem}
\usepackage{subcaption}
\usepackage[normalem]{ulem}
\useunder{\uline}{\ul}{}
\usepackage{hyperref}       

\title{Entropy Law: The Story Behind Data Compression and LLM Performance}

\author{%
Mingjia Yin$^\dagger$, Chuhan Wu$^{*^\dagger}$, Yufei Wang$^\dagger$, Hao Wang$^{*^\dagger}$,\\ \textbf{Wei Guo, Yasheng Wang, Yong Liu, Ruiming Tang$^*$, Defu Lian, Enhong Chen\thanks{Corresponding authors. $\dagger$ Equal contribution. } }\\
\small{State Key Laboratory of Cognitive Intelligence \& University of Science and Technology of China} \\
 \small{Noah's Ark Lab, Huawei}\\
\small{\{mingjia-yin, wanghao3, cheneh\}@ustc.edu.cn}\\ \small{ \{wuchuhan1, tangruiming\}@huawei.com} \\
}

\begin{document}

\maketitle

\begin{abstract}

\input{Content/0.abstract}
\end{abstract}

\section{Introduction}

\input{Content/1.introduction}

\section{Related Works}

\input{Content/2.related_work}



\section{Entropy Law: Connecting Model Performance with Data Compression}\label{sec: entropy_law}

\input{Content/3.5.theory}

\section{ZIP: Lightweight Data Selection for LLM Alignment}\label{sec: ZIP_algorithm}

\input{Content/4.method}

\section{Experiments}

\input{Content/5.experiment}

\section{Conclusion}

\input{Content/6.conclusion}


\medskip

\bibliographystyle{ACM-Reference-Format}
\bibliography{references}


\appendix

\input{Content/7.appendix}

\end{document}

%% file: Content/0.abstract.tex
Data is the cornerstone of large language models (LLMs), but not all data is useful for model learning.
Carefully selected data can better elicit the capabilities of LLMs with much less computational overhead.
Most methods concentrate on evaluating the quality of individual samples in data selection, while the combinatorial effects among samples are neglected.
Even if each sample is of perfect quality, their combinations may be suboptimal in teaching LLMs due to their intrinsic homogeneity or contradiction.
In this paper, we aim to uncover the underlying relationships between LLM performance and data selection.
Inspired by the information compression nature of LLMs, we uncover an ``entropy law'' that connects LLM performance with data compression ratio and first-epoch training loss, which reflect the information redundancy of a dataset and the mastery of inherent knowledge encoded in this dataset, respectively.
Through both theoretical deduction and empirical evaluation, we find that model performance is negatively correlated to the compression ratio of training data, which usually yields a lower training loss.
Based on the findings of the entropy law, we propose a quite efficient and universal data selection method named \textbf{ZIP} for training LLMs, which aim to prioritize data subsets exhibiting a low compression ratio.
Based on a multi-stage algorithm that selects diverse data in a greedy manner, we can obtain a good data subset with satisfactory diversity.
Extensive experiments have been conducted to validate the entropy law and the superiority of ZIP across different LLM backbones and alignment stages.
We also present an interesting application of entropy law that can detect potential performance risks at the beginning of model training.\footnote{Code can be found in \url{https://github.com/USTC-StarTeam/ZIP}.}

%% file: Content/1.introduction.tex
In recent years, Large Language Models (LLMs) have gained significant attention from both academia and industry, applied in various domains, such as chatbots~\citep{GPT3.5, GPT4}, chemistry tools~\citep{llm_application_chemistry}, and programming assistants~\citep{llm_application_github_copilot}.
The great success of LLMs depends on their general intelligence obtained from a vast amount of data collected from various sources~\citep{survey_data_selection, survey_llm_data_huawei}.
Through pretraining on trillions of tokens to master diverse knowledge and tuning on smaller instruction data to align models with human preference, LLMs can effectively utilize their knowledge to follow user instructions, do commonsense reasoning, and solve real-world problems~\citep{survey_LLM1}.

However, not all data are useful for teaching LLMs, especially when computational resources are limited~\citep{survey_data_selection}.
For example, we can better elicit the capability of LLMs by fine-tuning them on carefully curated samples rather than a large but noisy data collection~\citep{GPT3.5, PALM, Llama3, lima}.
However, selecting the proper data for LLM training is quite complicated and abstruse, since the space of data preprocessing and combination is almost unlimited.
Due to the huge computational overhead of LLM training, manual or empirical data selection based on trial-and-error feedback is rather cumbersome and even impractical.
Therefore, automatic data selection methods are necessary for LLM development under limited computational budgets.

\begin{figure*}
    \centering
    \includegraphics[width=\linewidth]{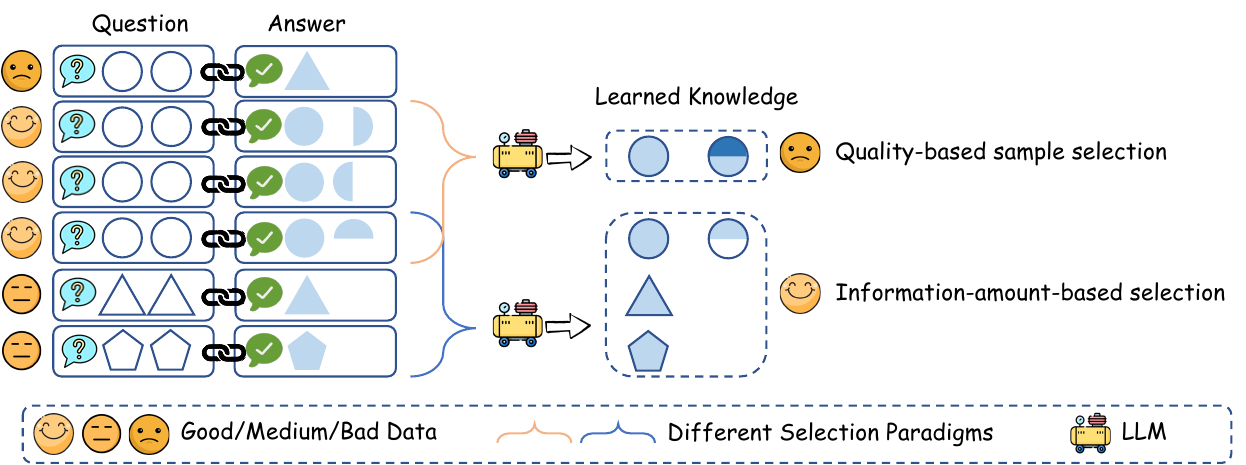}
    \caption{
    An illustrative example describing different data selection paradigms.
    Quality-based data selection relies on sample-level data quality measurements while overlooking combinatorial effects among samples.
   Information-amount-based selection aims to select samples maximizing the overall information amount.}
    \label{fig: motivation}
\end{figure*}

Intuitively, high-quality samples are expected to have better efficiency in teaching LLMs.
For example, the successful practice of LIMA~\citep{lima} shows the powerful effect of data quality on LLM performance that can surpass the amount of data.
Therefore, existing methods usually focus on quality-oriented data selection, based either on heuristic rules~\citep{english_only_c4, Gopher, DSIR, PALM, IFD} or evaluation models~\citep{QuRating, alpagasus, InsTag, Deita, ultrafeedback}.
Heuristic methods typically involve hand-crafted rules (e.g., sentence number~\citep{english_only_c4}, word count~\citep{Gopher}, length~\citep{longest}) to evaluate data across multiple dimensions.
Model-based approaches, on the contrary, rely on well-established LLMs such as GPT-4~\citep{GPT4} to provide quality assessments of training samples in different views, such as direct scoring~\citep{alpagasus}, task tagging~\citep{InsTag}, and pairwise scoring~\citep{Deita}.
However, most of these approaches evaluate different data samples independently, which neglects the intricate combinatorial effects among samples.
As illustrated in Figure~\ref{fig: motivation}, even if each sample is in perfect quality, their combinations may still be suboptimal due to their mutual information redundancy or inconsistency.
Although the quality-based subset is composed of all three good samples, the knowledge they encode is actually redundant and conflicting.
In contrast, another data subset composed of several relatively lower-quality but diverse samples may convey more information than the above subset in the teaching of LLMs.
Therefore, quality-based data selection does not fully align with the goal of maximizing the knowledge mastery of LLMs.

In many recent studies, researchers have shown that the basic mechanism of auto-regressive language modeling in LLMs is information compression~\citep{lm_is_compression, compression_represents_intelligence}.
Thus, the knowledge condensed by LLMs actually depends on the effective information encoded by training data.
This intuition opens another direction of data selection, i.e., based on the effective information amount of data.
In this paper, we uncover the underlying relations between LLM performance and data homogeneity, which can be measured by various canonical lossless compression algorithms (e.g., DEFLATE in ZIP).
Through both theoretical analysis and empirical experiments, we formulate the ``\textit{entropy law}'', which shows that the compression ratio of training data is a decisive factor affecting model performance, if the overall quality and consistency of selected samples remain unchanged.
Motivated by the entropy law, we propose an effective and efficient data selection algorithm called \textbf{ZIP} to select heterogeneous data with low compression ratio, which aims to maximize the effective information amount of information for LLM learning.
Specifically, we devise a multi-stage greedy strategy to find an approximate solution that guarantees a low compression ratio without exhausting all possible combinations, and it iterates continuously until we obtain a predetermined number of samples.
In each iteration, ZIP performs preliminary filtering to choose a smaller pool of candidates, and then selects a few samples from the reduced pool that minimizes the compression ratio of the selected dataset through a cascaded manner.
By learning LLMs on a collection of diverse samples that encode heterogeneous and complementary information, the capabilities of LLMs can be better elicited.
Extensive experiments on different LLM backbones at different stages of LLM alignment demonstrate the superiority of ZIP over various quality-based baselines.
We also present an interesting application of the entropy law that can detect potential performance risks at the beginning of model training, which can effectively reduce the computational overhead in LLM development.

%% file: Content/2.related_work.tex
\subsection{Large Modeling and Information Compression}

The relationship between language modeling and data compression has long intrigued researchers~\citep{Shannon_1, Shannon_2}.
\citet{gzip_scaling_law} has identified a data-dependant scaling law that takes data's gzip compressibility into consideration.
Besides, recent empirical studies have confirmed that language models can act as versatile data compressors~\citep{lm_is_compression}, and the intelligence of LLMs can be quantified by their capacity for text compression~\citep{compression_represents_intelligence}.
Let a text corpus be generated from an underlying distribution \( \rho \). A lossless compression algorithm \( \mathcal{C} \) is then expected to encode a text sequence \( x_{1:n} \) into a bitstream \( \mathcal{C}(x_{1:n}) \) of minimal length, ensuring that \( x_{1:n} \) can be perfectly recovered from \( \mathcal{C}(x_{1:n}) \). The expected number of bits of an optimal \( \mathcal{C} \) is equal to \( 
\mathbb{E}_{x \sim \rho} [-\log_2 \rho (x)] = \mathbb{E}_{x \sim \rho} [-\sum_{i=1}^{n} \log_2 \rho (x_i|x_{1:i-1})] \)~\citep{Shannon_1}.
The underlying distribution \( \rho \) is usually unknown in reality, but it can be estimated by a language model \( \rho_{\text{model}} \). Then the expected number of bits of an optimal \( \mathcal{C} \) can be updated:
\begin{equation}\label{eq: compression_is_cross_entropy}
\mathbb{E}_{x \sim \rho} [-\sum_{i=1}^{n} \log_2 \rho_{\text{model}} (x_i|x_{1:i-1})].
\end{equation}
Equation \ref{eq: compression_is_cross_entropy} is the cross-entropy loss employed in training LLMs, thereby establishing a coherent relationship between LLMs and information compression.
This foundational insight paves the way for this work.

\subsection{Alignment of Large Language Models} 

Large Language Models (LLMs) have recently gained significant attention from academia and industry.
LLM alignment, which includes supervised fine-tuning (SFT) and reinforcement learning with human feedback (RLHF), has emerged as a crucial technique for adapting LLMs to end tasks using natural language instructions~\citep{survey_LLM1, survey_llm_data_huawei}.
Alignment is performed using instruction datasets consisting of multiple (Instruction, Output) pairs, which require LLMs to follow the instructions and generate corresponding outputs.
Early explorations have focused on constructing or expanding instruction datasets through methods such as crowd-sourcing~\citep{sft_data_by_crowd_source1, sft_data_by_crowd_source2}, self-instruction~\citep{alpaca, alpaca-gpt4, self-instruct}, or the combination of existing datasets~\citep{tulu_v1, tulu_v2}.
Fine-tuned LLMs on these datasets have demonstrated promising capabilities to adhere to instructions across various contexts and align with human expectations.

\subsection{Data selection for LLM alignment}  
A growing body of research has emphasized the importance of selecting appropriate data for LLM alignment, which can prevent potential quality issues and optimize computational resource allocation.
As a prominent example, Lima~\citep{lima} has demonstrated superior performance by carefully crafting only 1,000 high-quality samples for SFT, highlighting the crucial importance of data quality.
The current literature on selecting alignment data has focused on selecting samples according to individual sample quality, which can be categorized into heuristic methods~\citep{longest} and model-based methods~\citep{alpagasus, InsTag, Deita, IFD, IFD2, MoDS}.
Heuristic methods typically employ specific criteria, such as response length~\citep{longest}, to guide data selection.
On the other hand, model-based methods adopt various strategies to leverage the capabilities of established language models for evaluating sample quality.
For example, IFD~\citep{IFD} measures the change in response loss when instructions are removed, and selects those with the most significant changes.
Building upon IFD, SuperFiltering~\citep{IFD2} introduces a lightweight proxy model for a more efficient calculation of the IFD score.
In addition, other model-based methods employ proprietary LLMs to assess data quality.
In a pioneering work, AlpaGasus~\citep{alpagasus} uses ChatGPT directly to assign data quality scores to samples, while \#InsTag~\citep{InsTag} proposes assigning tags to each sample using ChatGPT and evaluates sample quality based on the number of tags.
DEITA~\citep{Deita} uses ChatGPT-generated data to train two Llama-based scorers, assigning complexity and quality scores to each sample, and ultimately selecting samples with the highest hybrid scores.
However, existing methods are mainly designed to pick data based on sample-wise quality measurements, which are usually weak in reflecting the overall dataset quality.
In this paper, we focus on the relation between performance and dataset quality, which can be efficiently measured by data compression metrics.

%% file: Content/3.5.theory.tex
In this section, we provide some theoretical analysis of the relations between data compression and LLM performance.
Intuitively, the correctness and diversity of the training data would affect the performance of the final model.
Meanwhile, the performance of LLM may be suboptimal if the data have severe intrinsic conflicts or the model has poor mastery of the information encoded by the data, which can be indicated by the training loss.
Based on these assumptions, we denote the performance of an LLM as $Z$, which is expected to be influenced by the following factors:
\begin{itemize}[leftmargin=7.5mm]
    \item Data compression ratio $R$: This metric can be derived by dividing the pre-compression data size by the post-compression size, which can be computed by various off-the-shelf compression algorithms. Intuitively, a dataset with a lower compression ratio indicates a higher information density. 
    \item Training loss $L$: Indicates whether the data are hard for the model to memorize. Given the same base model, a high training loss is usually due to noisy or inconsistent information in the dataset. In practice, the average loss of a small number of training steps in the first training epoch is sufficient to produce an indicative $L$ value so that the model does not overfit the data.
    \item Data consistency $C$: The consistency of data is reflected by the entropy of the probability of the next token given the previous contexts. Higher data consistency usually yields a lower training loss.
    The performance of LLMs is usually suboptimal if the dataset has poor consistency.\footnote{
    Assume we have two question-answer pairs $(q_1, a_1)$ and $(q_2, a_2)$. When an LLM updates on each QA pair, it learns the mutual information (MI) between the question and answer, i.e., $I(q_1; a_1)$ and $I(q_2; a_2)$. When it updates on both, it learns the joint MI $I(q_1q_2; a_1a_2)$. If $q_1$ and $q_2$ are not independent, we have $I(q_1q_2; a_1a_2) < I(q_1; a_1) + I(q_2; a_2)$, whose detailed derivation can be found in Appendix \ref{sec: mutual_information_derivation}. This implies that the total knowledge learned by LLMs is narrowed if the answers to similar questions are highly inconsistent.}
    \item Average data quality $Q$: This reflects the average sample-level quality of the data, which can be measured through various objective and subjective aspects.
\end{itemize}
Given a certain amount of training data, the model performance can be estimated by the above factors:
\begin{equation}\label{eq: original_model_performance}
    Z \propto f(R, L, C, Q),
\end{equation}
where $f$ is a hidden function.
Given a specific base model, the scale of $L$ usually depends on $R$ and $C$, which can be formulated as:
\begin{equation}\label{loss,r,c}
    L \propto g(R, C).
\end{equation}
$L$ is expected to be monotonous on $R$ and $C$, since a dataset with higher homogeneity or better data consistency is easier for a model to learn.
Thus, we can rewrite the above formula as follows:
\begin{equation}
    C \propto g'(R, L),
\end{equation}
where $g'$ is an inverse function.
By combining the three above equations, we have:
\begin{equation}
    Z  \propto  f(R, L, g'(R, L), Q)  \propto  h(R, L, Q),
\end{equation}
where $h$ is another hidden function.
If a data selection method does not substantially change the average data quality $Q$, we can approximately regard the variable $Q$ as a constant.
Therefore, the final performance can be roughly formulated as follows:
\begin{equation}
    Z \propto h'(R, L),
\end{equation}
which means that the model performance is correlated with the data compression ratio and training loss.
We name this relationship as ``\textbf{Entropy Law}''.

We can raise two deductions based on the entropy law:
\begin{itemize}[leftmargin=7.5mm]
    \item If we further regard the data consistency as a constant, the training loss is directly influenced by the compression ratio (Eq. \ref{loss,r,c}). Thus, the model performance is controlled by the compression ratio: $Z$ is usually worse if the data compression ratio $R$ is higher, which will be validated by our experiments.
\item Given the same compression ratio $R$, a higher training loss means a lower data consistency. Thus, the effective knowledge learned by the model may be more limited. This can be used to predict the performance of LLM on different data with similar compression ratios and sample qualities. We will show later the application of this deduction in our practice.
\end{itemize}

Notably, entropy law reveals a coherent connection between downstream model performance and data compression ratio, setting it apart from the previously proposed data-dependent scaling law by \citet{gzip_scaling_law}.
Building upon the entropy law, we derive a data selection algorithm in Section \ref{sec: ZIP_algorithm} and demonstrate its application in practical large-scale LLM development in Section \ref{sec: exp_entropy_law}.

%% file: Content/4.method.tex
Guided by the findings of the entropy law, we propose an effective and efficient method named \textbf{ZIP} to select data samples based on data compression ratios, which aims to maximize the amount of effective information given a limited training data budget.
Although there exists a subset with the lowest compression ratio, it is impractical to find it due to the huge combination space of data samples.
Thus, we propose an iterative multi-stage greedy algorithm to efficiently obtain an approximate solution with a relatively low compression ratio.
In each iteration, we first use a global selection stage to choose a pool of candidate samples that have low compression ratios, which aims to find samples with high information density.
We then employ a coarse-grained local selection stage incorporating a smaller set of samples with the lowest redundancy with already selected samples.
Finally, we use a fine-grained local selection stage that minimizes the similarity between samples to add.
The above process is conducted until we obtain a sufficient size of data.
The workflow of our method is summarized in Algorithm~\ref{alg: data_selection}, whose details are introduced as follows.

\subsection{Global Selection}

In general, we maintain an information redundancy state $\pi_{\mathcal{D}}$ that evaluates the ``information gain'' of each sample.
Intuitively, data with high intra-sample information redundancy are unlikely to have good global diversity.
For example, a sample with repeated patterns or echoed conversation turns usually has low education value in LLM training.
Thus, we initialize this state by calculating the sample-level compression ratio for the entire dataset $\mathcal{D}$. 
In each iteration, We select $K_1$ samples with the lowest scores in $\pi_{\mathcal{D}}$ to form an initial candidate pool $\mathcal{D}_{K_1}$, which provides a good set for subsequent local selection.

\subsection{Local Coarse-grained Selection}

Since the global selection does not well consider the mutual relations among samples, we further conduct local selection to pick diverse samples.
To ensure good computational efficiency, we introduce a coarse-grained selection phase to narrow the candidate pool into a smaller one with $K_2$ samples.
We first compute the compression ratio of a merged set that adds each sample in $\mathcal{D}_{K_1}$ to the selected set $\mathcal{D}'$.
We use this score to update the information redundancy state $\pi_{\mathcal{D}}$ to better indicate the current information gain of these samples.
Based on the scores of the samples in $\mathcal{D}_{K_1}$, we select $K_2$ samples with the lowest scores.
These samples form a small subset for final fine-grained selection, where each sample has good diversity with the selected dataset $\mathcal{D}'$.

\input{Algorithm/data_selection}

\subsection{Local Fine-grained Selection}

Although the above stage ensures that the candidate pool has distinct information from the selected set, the information redundancy among the samples within this pool is not measured.
Thus, we aim to pick further samples from this subset that are diverse from each other.
Concretely, we initialize a local selected set $\mathcal{D}_{K_3}$, and compute the compression ratio of the union of $\mathcal{D}_{K_3}$ and each sample in $\mathcal{D}_{K_2}$.
We add the sample with the lowest compression ratio into $\mathcal{D}_{K_3}$, and remove it from $\mathcal{D}_{K_2}$.
By repeating this process, we obtain a small subset that contains samples not only different from the selected set $\mathcal{D}'$ but also distinct from each other.
We conduct the three stages above until the size of $\mathcal{D}'$ reaches a predefined data budget $m$.
In our method, the entire selection process is quite easy to implement since it is model-free, and can be accelerated using multiple threads.
It can select data efficiently and effectively from a large candidate pool for high-quality LLM training.

%% file: Algorithm/data_selection.tex
\begin{algorithm}[t]
\renewcommand{\algorithmicrequire}{\textbf{Input:}}
\renewcommand{\algorithmicensure}{\textbf{Output:}}
\caption{Pseudo code of ZIP}
\label{alg: data_selection}
\begin{algorithmic}[1]
\Require The original dataset $\mathcal{D}$ of size $N$. A data compressor $\mathcal{C}$. The number of selected samples $m$. The number of compression ratios to be updated in the global selection $K_1$. The data pool size of local selection $K_2$. The data budget of local selection $K_3$. Compression ratio calculation function $g(\mathcal{C}(D))=\frac{\text{Bits}(D)}{\text{Bits}(\mathcal{C}(D))}$, which measures compression ratio by the number of bits.
\Ensure A data subset $\mathcal{D}'$ of size $m$ with a relatively low compression ratio.

\State Init $\mathcal{D}'$ as an empty set $\Phi$
\State Init $\pi_{\mathcal{D}}=\{g(d_0), g(d_1), \dots, g(d_N)\}$ by calculating the compression ratio of each sample in $\mathcal{D}$

\While{$|\mathcal{D}'| < m$}
    \State // \textbf{Stage 1: Global Selection}
    \State $\mathcal{D}_{K_1}$ = Bottom-K($\mathcal{D} \setminus \mathcal{D}'$, $\pi_{\mathcal{D} \setminus \mathcal{D}'}$, $K_1$) // Select $K_1$ samples with the lowest scores 
    \State // \textbf{Stage 2: Local Coarse-grained Selection}
    \State $\pi_{K_1} = g(\mathcal{D}' \cup \{ d \})$, where $d \in \mathcal{D}_{K_1}$ //Compression ratios of the union of $\mathcal{D}'$ and $d$
    \State Update corresponding values in $\pi_{\mathcal{D}}$ with $\pi_{K_1}$
    \State $\mathcal{D}_{K_2}$ = Bottom-K($\mathcal{D}_{K_1}$, $\pi_{K_1}$, $K_2$) // Select $K_2$ samples with the lowest compression ratios
    \State // \textbf{Stage 3: Local Fine-grained Selection}
    \State Init $\mathcal{D}_{K_3}$ as an empty set $\Phi$
    \While{$|\mathcal{D}_{K_3}| < K_3$}
        \State $\pi_{K_2} = \{g(\mathcal{D}_{K_3} \cup \{ d \}) | d \in \mathcal{D}_{K_2} \}$
        \State d = $\argmin_{d}{\pi_{K_2}}$
        \State $\mathcal{D}_{K_3}$ = $\mathcal{D}_{K_3} \cup \{d\}$
        \State $\mathcal{D}_{K_2}$ = $\mathcal{D}_{K_2} \setminus \{d\}$
    \EndWhile
    \State // \textbf{Update the selected dataset}
    \State $\mathcal{D}' = \mathcal{D}' \cup \mathcal{D}_{K_3}$

\EndWhile

\end{algorithmic}
\end{algorithm}

%% file: Content/5.experiment.tex
ZIP is content-agnostic and model-free, making it suitable for various stages of LLM alignment.
We systematically evaluate the effectiveness of ZIP through experiments conducted in the SFT and RLHF stages, as described in Sections \ref{sec: exp_sft} and \ref{sec: exp_rlhf}, respectively.
Subsequently, Section \ref{sec: exp_entropy_law} presents an in-depth analysis to empirically support the proposed entropy law, including a practical application guided by this law.

\subsection{Data Selection for SFT}\label{sec: exp_sft}

\subsubsection{Setup}\label{sec: sft_setup}

\textbf{Data Pool \& Data Selection} \space
We follow DEITA~\citep{Deita} to establish a large-scale data pool comprising 300K high-quality samples obtained from WizardLM~\citep{WizardLM}, ShareGPT~\citep{sharegpt}, and UltraChat~\citep{ultrachat}.
Subsequently, various data selection techniques are employed to extract a subset of this pool for LLM instruction tuning. 
Notably, previous studies controlled the data budget by limiting the number of instances, whereas we managed the total token count to ensure a fair allocation of the compute budget among all methods.
To achieve this, we initially select 10,000 samples using ZIP and calculate the corresponding token count.
Then, we apply other methods to continue data selection until the required token count is reached.

\textbf{Training \& Evaluation} \space
We fine-tune Mistral-7B~\citep{mistral} and LLama-3-8B~\citep{Llama3} on the selected dataset.
Other training details can be found in Appendix \ref{sec: training_details}.
As for evaluation, we adopt MT-bench\citep{mt-bench} as our benchmark.
Specifically, MT-bench is a challenging multi-turn question set with LLM judgements to evaluate model responses, which exhibits a high-level human preferences alignment.

\textbf{Baselines} \space
We select the baseline from two aspects. The first group includes heuristic methods: (1) \textit{Random}, which randomly selects instances from the data pool to verify the fundamental effectiveness of other methods; (2) \textit{Cluster}, which adopts K-means clustering based on the sample representations and select cluster centroids; (3) \textit{Perplexity}, which selects the samples with highest training loss.
The second group of baselines includes model-based methods: (1) \textit{DEITA}~\citep{Deita}, which employs ChatGPT-generated data to train a Llama-based data complexity evaluator and a quality evaluator, and selects samples with the highest hybrid scores; (2) \textit{SuperFiltering}~\citep{IFD2}, which assesses each sample by calculating the change in response loss upon instruction removal and introduce a lightweight proxy model to calculate the score more efficiently.

\subsubsection{Results}
\input{Table/SFT_result}

\textbf{Main comparison} \space
We compare ZIP with various data selection methods based on Mistral-7B and Llama-3-8B, and the results are presented in Table \ref{tab: SFT_result}.
ZIP outperforms other data selection approaches on all backbones, which can be attributed to ZIP's ability to model the complex combinatorial effects among samples.
Furthermore, our observations indicate that model-based data selection methods often fail to produce satisfactory outcomes when a fixed token number is given.
This is because the sample-level evaluations are not updated correspondingly after selecting some samples, leading to biased evaluations for the remaining samples.
Additionally, some of these methods adopt strategies to enhance data diversity, such as \textit{DEITA}, which controls the representation distances of selected samples.
However, these strategies only provide a rough assessment of the combinatorial effects within the representation space, since semantic distances do not necessarily reflect information redundancy.

\textbf{Selection bias in sample length across different strategies} \space
We also provide the average length of tokenized samples in Table \ref{tab: SFT_result}.
The average token length of \textit{Random} provides an estimation for the entire data pool, which is used to analyze other methods.
From the tables, we can observe that \textit{Cluster} and \textit{Perplexity} exhibit similar selection preferences as \textit{Random}.
Additionally, \textit{Deita} and \textit{SuperFiltering} predominantly select lengthy data samples.
This bias may stem from the LLMs' inclination toward generating longer responses~\citep{length_bias}.
However, given the limited budget of selected tokens, choosing excessively lengthy data will reduce the information density and degrade the capabilities of models trained on such data.
In contrast, ZIP tends to select shorter samples.
Furthermore, we plot the token length distribution of these methods, as depicted in Figure \ref{fig: mistral_distribution} and Figure \ref{fig: llama3_distribution}.
Consistent with the previous results, we can observe similar distributions for \textit{Random}, \textit{Cluster}, and \textit{Perplexity}.
The token length distributions of \textit{DEITA} and \textit{SuperFiltering} are severely skewed, deviating greatly from the original data distribution.
In contrast to these model-based approaches, ZIP exhibits no bias toward selecting lengthy samples.

\textbf{Cost comparison of different strategies} \space
We provide a detailed cost analysis of each method in Table \ref{tab: SFT_result}.
Except for the \textit{Random} method, ZIP required the least time to complete the data selection process, demonstrating greater efficiency than other methods.
Notably, ZIP's computations are entirely executed on CPUs, resulting in significant cost savings.
Furthermore, ZIP is independent of proprietary LLMs used by \textit{DEITA} or the proxy model employed by \textit{Cluster}, \textit{Perplexity}, and \textit{SuperFiltering}.
This model-free characteristic endows ZIP with notable efficiency and versatility.

\input{Figure/mistral_distribution}

\textbf{Selected data quality of different strategies} \space
We have followed Alpagasus~\citep{alpagasus} to evaluate the quality of each data sample in the selected datasets by prompting ChatGPT, with the quality scores ranging from 0 to 5.
The quality scores of multi-turn samples are the average scores of each turn.
The results have been presented in Table \ref{tab: SFT_result}.
Surprisingly, the quality scores of selected datasets are highly similar, even with significant differences in selection mechanisms.
This may suggest that the average quality distribution remains relatively uniform in the original data pool.
Notably, even the SOTA model-based methods like \textit{DEITA}~\citep{Deita} and \textit{SuperFiltering}~\citep{IFD2} select data with similar quality scores, potentially contradicting their original conclusions.
We posit that this discrepancy stems from the setting of the data budget, which is controlled by the number of samples in prior studies.
Considering the selection bias discussed above, these methods tend to select lengthy samples, resulting in a significantly higher token count compared with baselines.
For instance, under this setting, data selected by \textit{DEITA} will possess 2.7 times the number of tokens compared to \textit{ZIP}.
However, we argue it is fairer to control the data budget by the token count since it guarantees a similar compute budget among all methods\footnote{In practical implementation, the training steps of all methods are almost equal by employing the packing technique detailed in \href{https://github.com/OpenAccess-AI-Collective/axolotl/blob/main/docs/multipack.qmd}{Axolotl}.}.

\subsection{Data Selection for RLHF}\label{sec: exp_rlhf}
\input{Table/RLHF_llama3}

\subsubsection{Setup}

\textbf{Data Pool \& Data Selection} \space
The data pool used for preference alignment is a cleaned version of UltraFeedback~\citep{ultrafeedback, Cleaned_ultra_feedback}, which consists of around 60k samples in the form of a "chosen-rejected" pair.
Similarly to the SFT stage, we ensure each data selection method selects data with an approximately equal token count.
Since a "chosen-rejected" data pair encompasses two data points, we select 5,000 data pairs with ZIP and then apply other methods to select data with the corresponding token budget.
\input{Figure/entropy-law-mistral}
\input{Figure/entropy-law-llama3}

\textbf{Training \& Evaluation} \space
Building upon the model previously fine-tuned with SFT, we further refine it using RLHF.
In particular, we employ Kahneman-Tversky Optimization (KTO)~\citep{KTO} for preference alignment, a novel method that shows promising potential in aligning preferences.
Additional training details can be found in Appendix \ref{sec: training_details}.
For evaluation, we continue to utilize MT-bench~\citep{mt-bench} as our benchmark to assess the capabilities of LLMs fine-tuned with data selected using diverse data selection strategies.

\textbf{Baselines} \space
We compare ZIP with the following baselines: (1) \textit{Random}, which randomly samples some "chosen-rejected" pairs from the data pool. (2) \textit{Score}, which selects the "chosen-rejected" pairs with the highest "chosen-scores".
These scores are obtained through LLM evaluation of the response quality~\citep{ultrafeedback, Cleaned_ultra_feedback}.

\subsubsection{Main results}
Table \ref{tab: RLHF_result_llama3} presents the results of different data selection strategies on the preference alignment stage of LLMs.
Similar to the SFT stage, models aligned with data selected by ZIP can yield the best downstream performance, demonstrating the necessity for modeling the combinatorial effects.
Besides, we find \textit{Score} and \textit{Random} are on par with each other, even though the selection process of \textit{Score} is far more expensive than \textit{Random}.
This is unsurprising, as \textit{Score} does not consider the combinatorial effects, which may limit the knowledge amount of the selected dataset.

\subsection{Empirical Validation of Entropy Law}\label{sec: exp_entropy_law}

\input{Figure/entropy_law_application}

In this section, we aim to demonstrate the proposed entropy law.
Specifically, we have plotted the model performance of Mistral-7B and Llama-3-8B concerning data compression ratio and training loss in Figure \ref{fig: entropy_law_mistral} and \ref{fig: entropy_law_llama3}, respectively.
Besides, we plot entropy-law curves by fitting the results.
From the two figures, we can draw the following analysis:

\textbf{Relationship between model performance, data compression ratio, and training loss} \space
In Figure \ref{fig: mistral_entropy_law_compress} and Figure \ref{fig: llama3_entropy_law_compress}, LLMs trained on data with a lower compression ratio typically exhibit enhanced performance.
Since the learning process of LLMs is highly relevant to information compression, we can regard LLMs as data compressors.
Then the data with a lower compression ratio means a higher knowledge amount, which is more beneficial to the compressors.
Besides, a lower compression ratio usually corresponds a higher training loss, as illustrated in Figures \ref{fig: mistral_entropy_law_loss} and \ref{fig: llama3_entropy_law_loss}.
This is because data resistant to compression carries more knowledge, posing a greater challenge for LLMs to absorb the encapsulated knowledge.

\textbf{Model performance interpretation with entropy law} \space
Considering the three methods with comparable compression ratios and training loss, namely \textit{Random}, \textit{Cluster}, and \textit{Perplexity}, the corresponding model performances are close.
This phenomenon may seem counter-intuitive, given the distinct criteria used for data selection.
However, it aligns with the predictions of our proposed entropy law: when the average data quality, training loss, and data compression ratio are similar, the model performance is expected to be comparable as well.
Thus, the entropy law has the potential to serve as a criterion for predicting the model's performance on data, thereby guiding the training of LLMs.

\textbf{Practical application of entropy law} \space
Incremental version update of training data is a common setting in practical LLM development.
Usually, the training data amount remains relatively stable, with only a minor portion undergoing modification.
We have conducted incremental training data update experiments in real scenarios, with results depicted in Figure \ref{fig: entropy_law_application}.
Due to confidentiality, only the relative order of the results is provided.
Guided by entropy law, assuming the data quality $Q$ does not significantly decay after each incremental update, we can expect a model performance improvement with a decreased data compression ratio.
This expectation is supported by the results from data versions $x_1$ to $x_4$ in Figure \ref{fig: entropy_law_application} since their compression ratios are decreased after each incremental update.
However, the data version $x_5$ exhibits an abnormal increase in the loss and data compression ratio, which serves as an early indicator of potential model performance degradation due to a decline in training data consistency.
This prediction is further confirmed by subsequent post-training model performance evaluations, as illustrated in Figure \ref{fig: entropy_law_application}.
Thus, the entropy Law can be utilized as a guideline for LLM training to identify potential risks of experimental failure without training the model on the full dataset until convergence.
This is particularly significant given the substantial costs associated with training an LLM.

%% file: Table/SFT_result.tex
\begin{table}[]
\centering
\caption{Model performance comparison between different data selection baselines based on Mistral-7B and Llama-3-8B on the SFT stage. We also provide the computational cost, average token length of selected data, and the average data quality produced by Alpagasus~\cite{alpagasus}. The best results are bolded, and the second-best numbers are underlined. $\dagger$ means CPU-only methods.}
\label{tab: SFT_result}
\begin{tabular}{lrrrr}
\toprule
Model            & MT-bench $\uparrow$ & Cost $\downarrow$ & Avg.length & Quality \\ \midrule
\multicolumn{5}{c}{Mistral-7B-based models with SFT}                             \\ \midrule
Random $\dagger$ & 6.85                & \textbf{10s}     & 976        & 4.08    \\
Cluster          & {\ul 6.91}          & 15h              & 970        & 4.05    \\
Perplexity       & 6.89                & 8h               & 981        & 4.09    \\
SuperFiltering   & 6.12                & 14h              & 1579       & 4.10    \\
DEITA            & 6.82                & 21h              & 2048       & 4.03    \\
ZIP $\dagger$    & \textbf{7.08}       & {\ul 4.5h}       & 543        & 4.00    \\ \midrule
\multicolumn{5}{c}{Llama-3-8B-based models with SFT}                             \\ \midrule
Random $\dagger$ & 7.16                & \textbf{10s}     & 892        & 4.08    \\
Cluster          & {\ul 7.18}          & 16h              & 886        & 3.95    \\
Perplexity       & 7.09                & 9h               & 895        & 3.96    \\
SuperFiltering   & 6.59                & 14h              & 1481       & 3.99    \\
DEITA            & 7.11                & 21h              & 2048       & 4.09    \\
ZIP $\dagger$    & \textbf{7.28}       & {\ul 4.5h}       & 470        & 4.00    \\ \bottomrule
\end{tabular}
\end{table}

%% file: Figure/mistral_distribution.tex
\begin{figure}[t]
    \centering
    \begin{subfigure}[b]{0.3\textwidth} 
        \includegraphics[width=\textwidth]{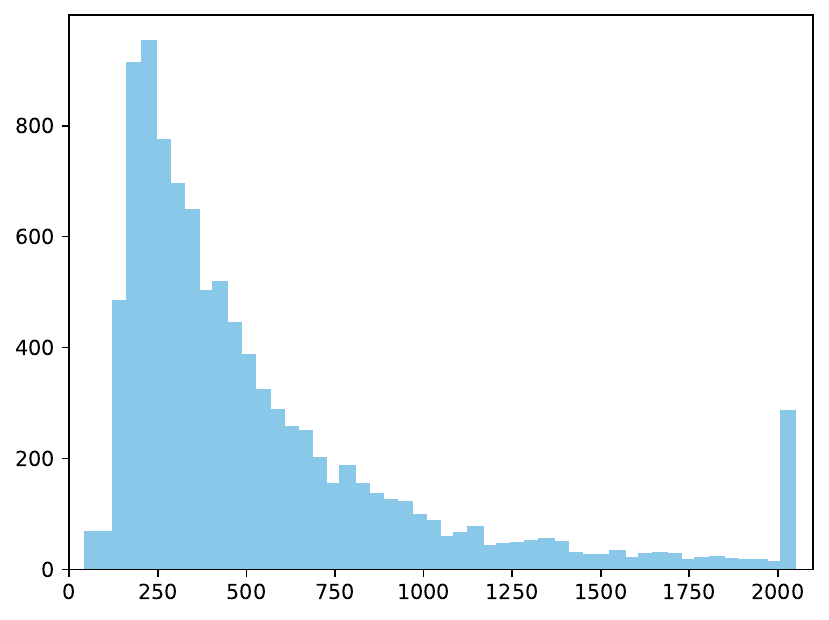}
        \caption{ZIP}
        \label{fig: mistral_distribution_zip}
    \end{subfigure}
    \hfill
    \begin{subfigure}[b]{0.3\textwidth} 
        \includegraphics[width=\textwidth]{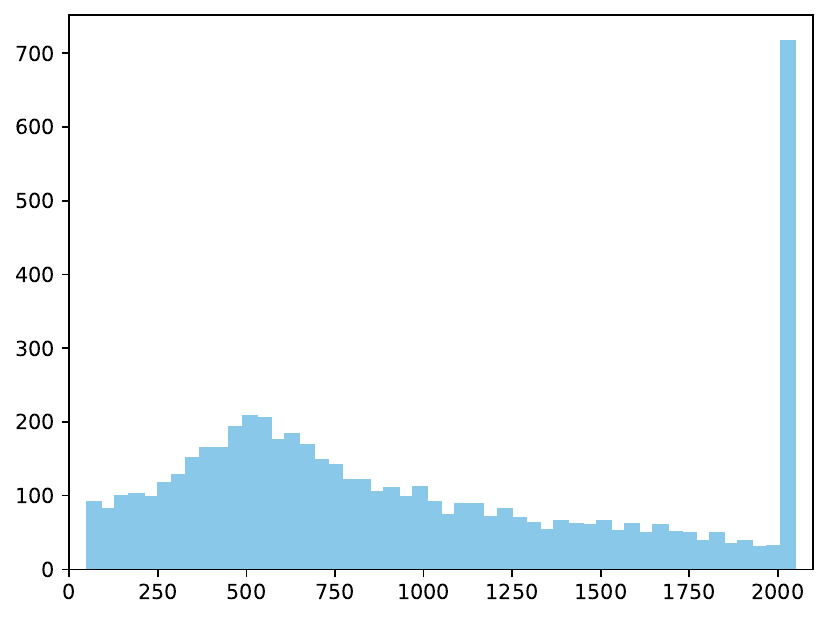}
        \caption{Random}
        \label{fig: mistral_distribution_random}
    \end{subfigure}
    \hfill
    \begin{subfigure}[b]{0.3\textwidth} 
        \includegraphics[width=\textwidth]{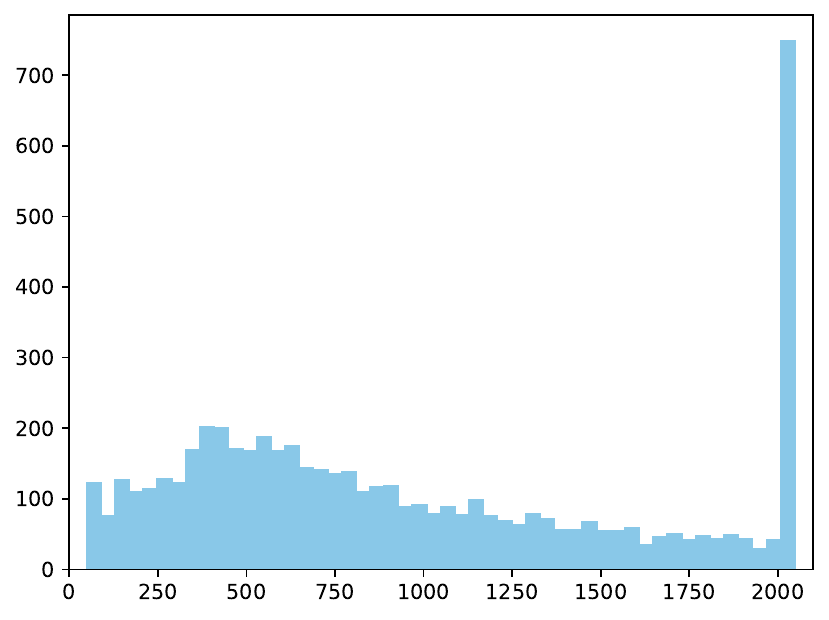}
        \caption{Diversity}
        \label{fig: mistral_distribution_diversity}
    \end{subfigure}

    \vfill 

    \begin{subfigure}[b]{0.3\textwidth} 
        \includegraphics[width=\textwidth]{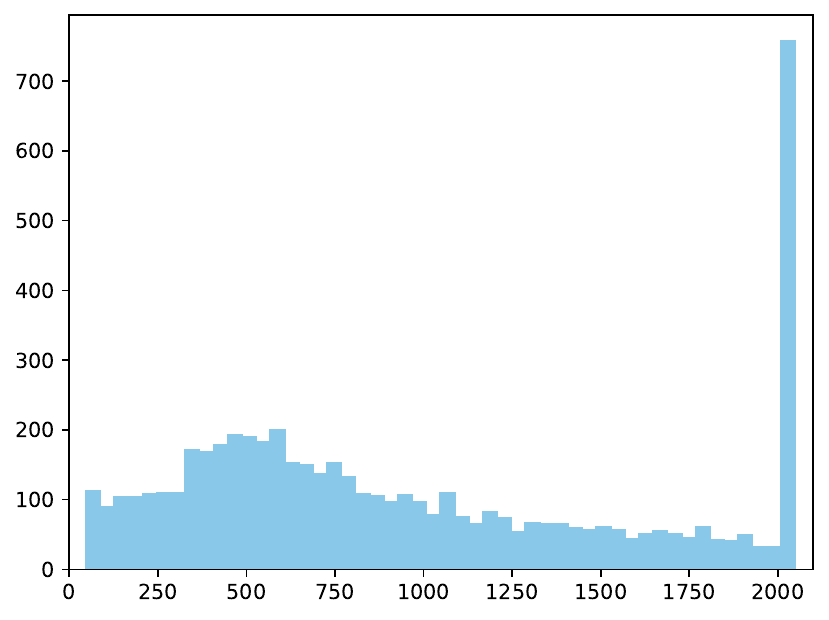}
        \caption{Perplexity}
        \label{fig: mistral_distribution_perplexity}
    \end{subfigure}
    \hfill
    \begin{subfigure}[b]{0.3\textwidth} 
        \includegraphics[width=\textwidth]{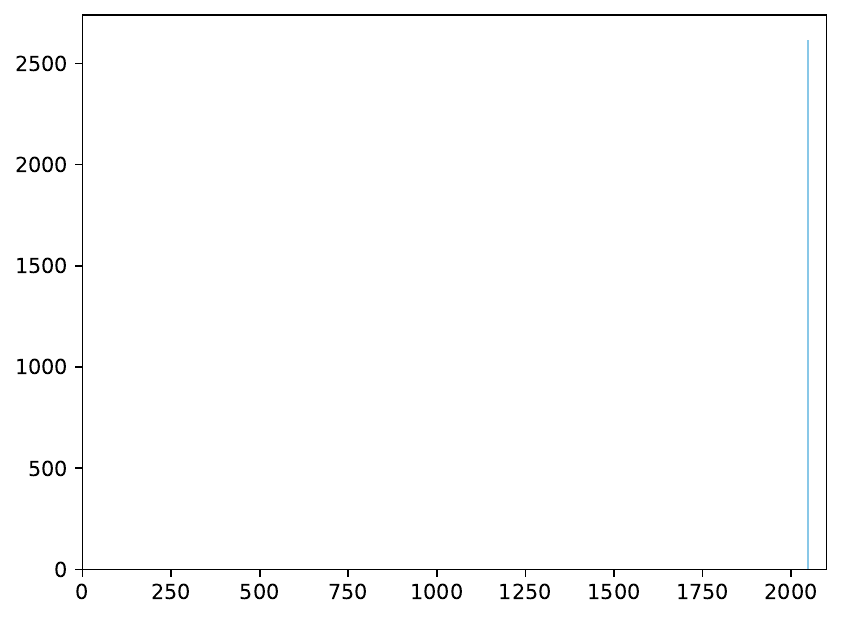}
        \caption{DEITA}
        \label{fig: mistral_distribution_deita}
    \end{subfigure}
    \hfill
    \begin{subfigure}[b]{0.3\textwidth} 
        \includegraphics[width=\textwidth]{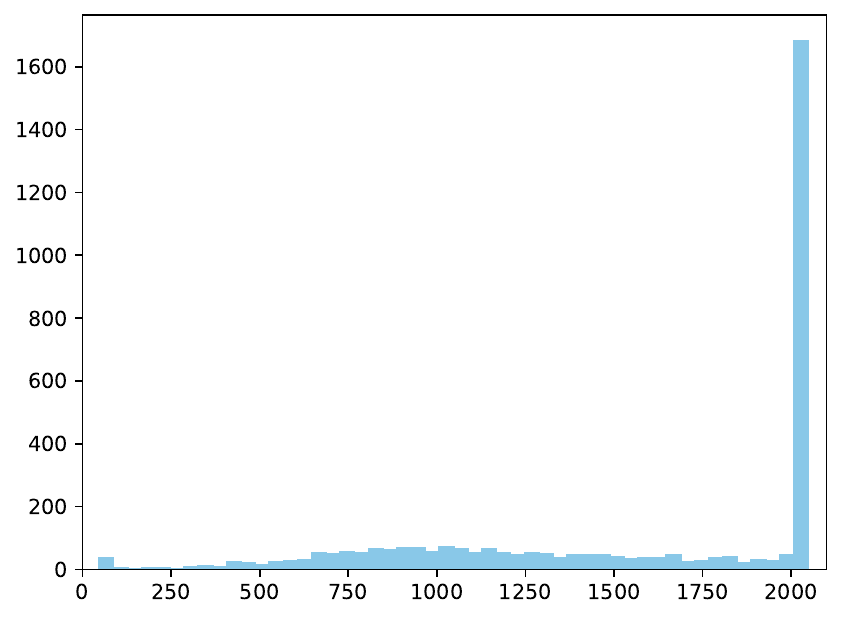}
        \caption{SuperFiltering}
        \label{fig: mistral_distribution_IFD}
    \end{subfigure}
    \caption{The distribution of average token number across datasets selected by different algorithms for Mistral-7B.}
    \label{fig: mistral_distribution}
\end{figure}

%% file: Table/RLHF_llama3.tex
\begin{table}[t]
\centering
\caption{Model performance comparison between different data selection baselines based on Llama-3-8B on the RLHF stage. We also provide the computational cost of each method.}
\label{tab: RLHF_result_llama3}
\begin{tabular}{lrrr}
\toprule
Model            & MT-bench $\uparrow$ & Cost & Avg.length \\ \midrule
Base             & 7.18                & NA   & NA         \\
Random $\dagger$ & {\ul 7.33}          & 5s   & 464        \\
Score            & 7.30                & NA   & 489        \\
ZIP $\dagger$    & \textbf{7.42}       & 1.1h & 357        \\ \bottomrule
\end{tabular}
\end{table}

%% file: Figure/entropy-law-mistral.tex
\begin{figure}[t]
    \centering
    \begin{subfigure}[b]{0.45\textwidth} 
        \includegraphics[width=\textwidth]{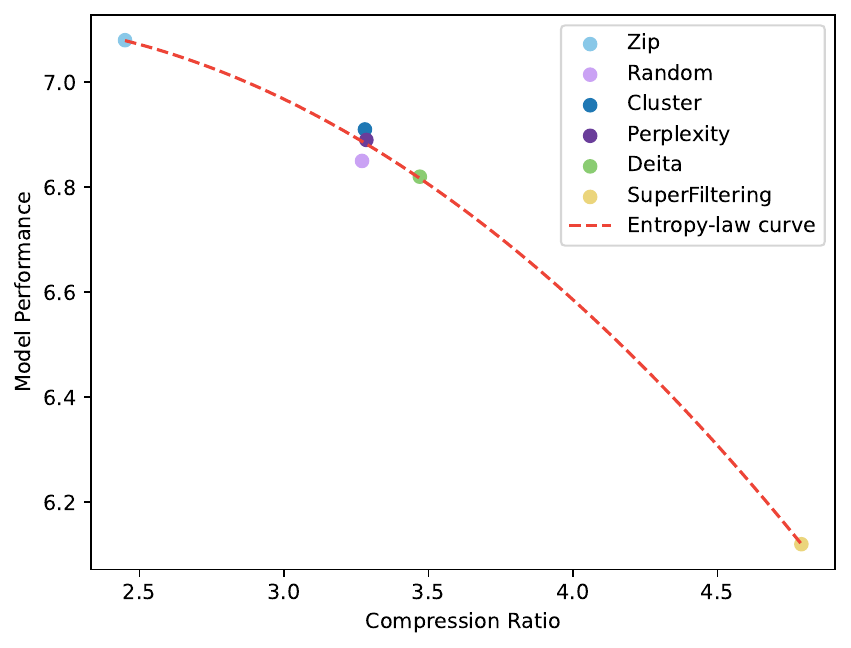}
        \caption{Entropy law w.r.t. compression ratio}
        \label{fig: mistral_entropy_law_compress}
    \end{subfigure}
    \hfill
    \begin{subfigure}[b]{0.45\textwidth} 
        \includegraphics[width=\textwidth]{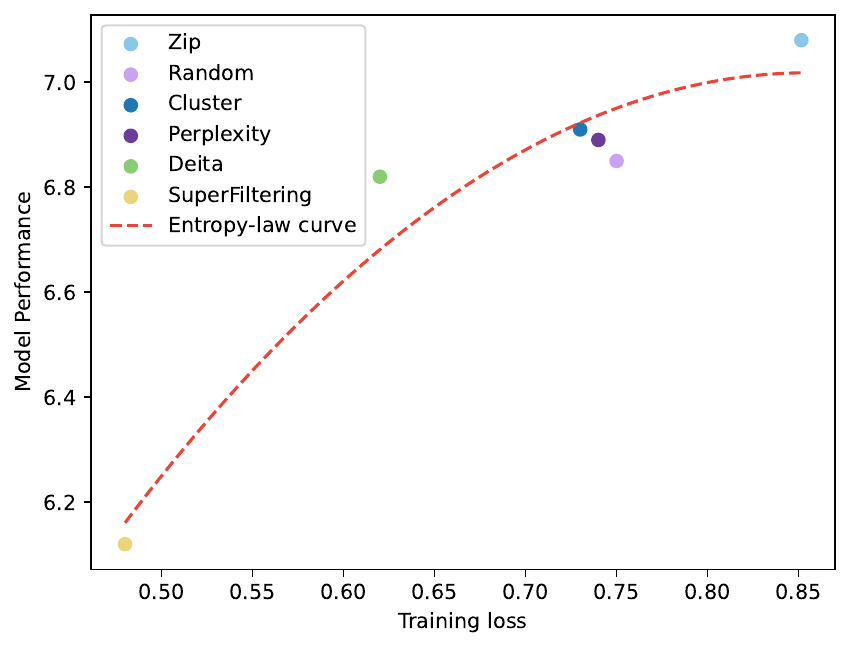}
        \caption{Entropy law w.r.t. training loss}
        \label{fig: mistral_entropy_law_loss}
    \end{subfigure}
    \caption{Entropy law demonstration of Mistral-7B. The Entropy law curve is fitted with the results of different methods.}
    \label{fig: entropy_law_mistral}
\end{figure}

%% file: Figure/entropy-law-llama3.tex
\begin{figure}[t]
    \centering
    \begin{subfigure}[b]{0.45\textwidth} 
        \includegraphics[width=\textwidth]{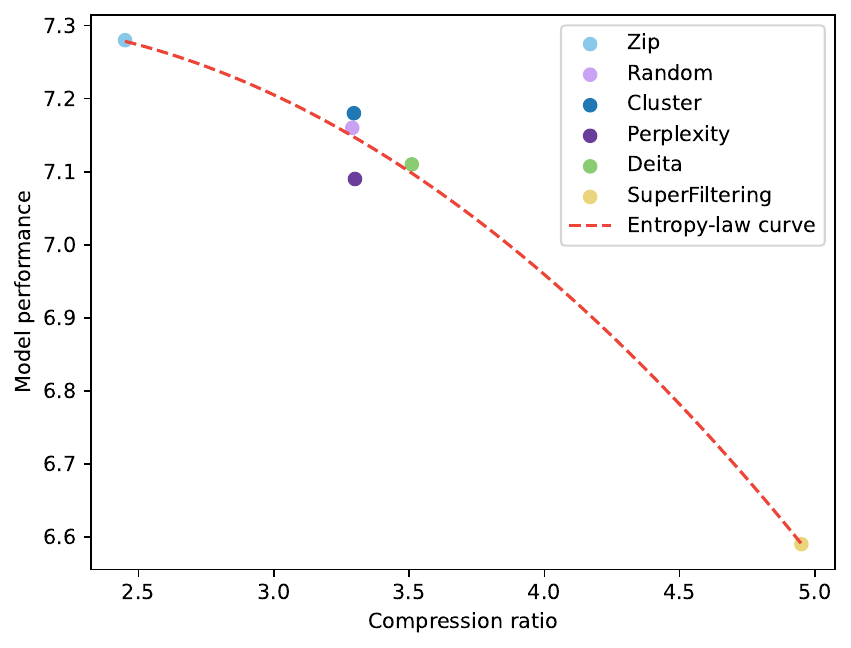}
        \caption{Entropy law w.r.t. compression ratio}
        \label{fig: llama3_entropy_law_compress}
    \end{subfigure}
    \hfill
    \begin{subfigure}[b]{0.45\textwidth} 
        \includegraphics[width=\textwidth]{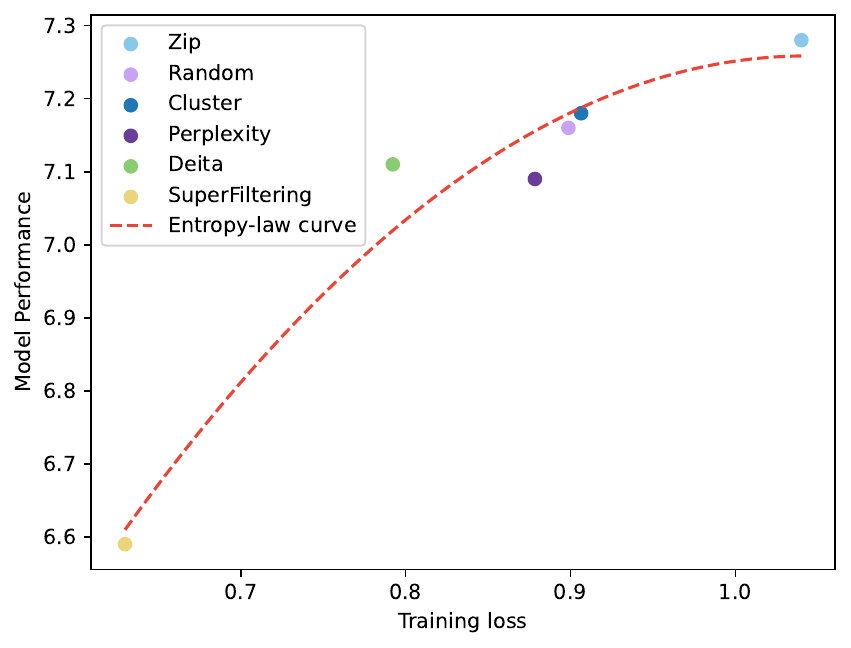}
        \caption{Entropy law w.r.t. training loss}
        \label{fig: llama3_entropy_law_loss}
    \end{subfigure}
    \caption{Entropy law curve of Llama-3-8B. The Entropy law curve is fitted with the results of different methods.}
    \label{fig: entropy_law_llama3}
\end{figure}

%% file: Figure/entropy_law_application.tex
\begin{figure}[t]
    \centering
    \includegraphics[scale=1]{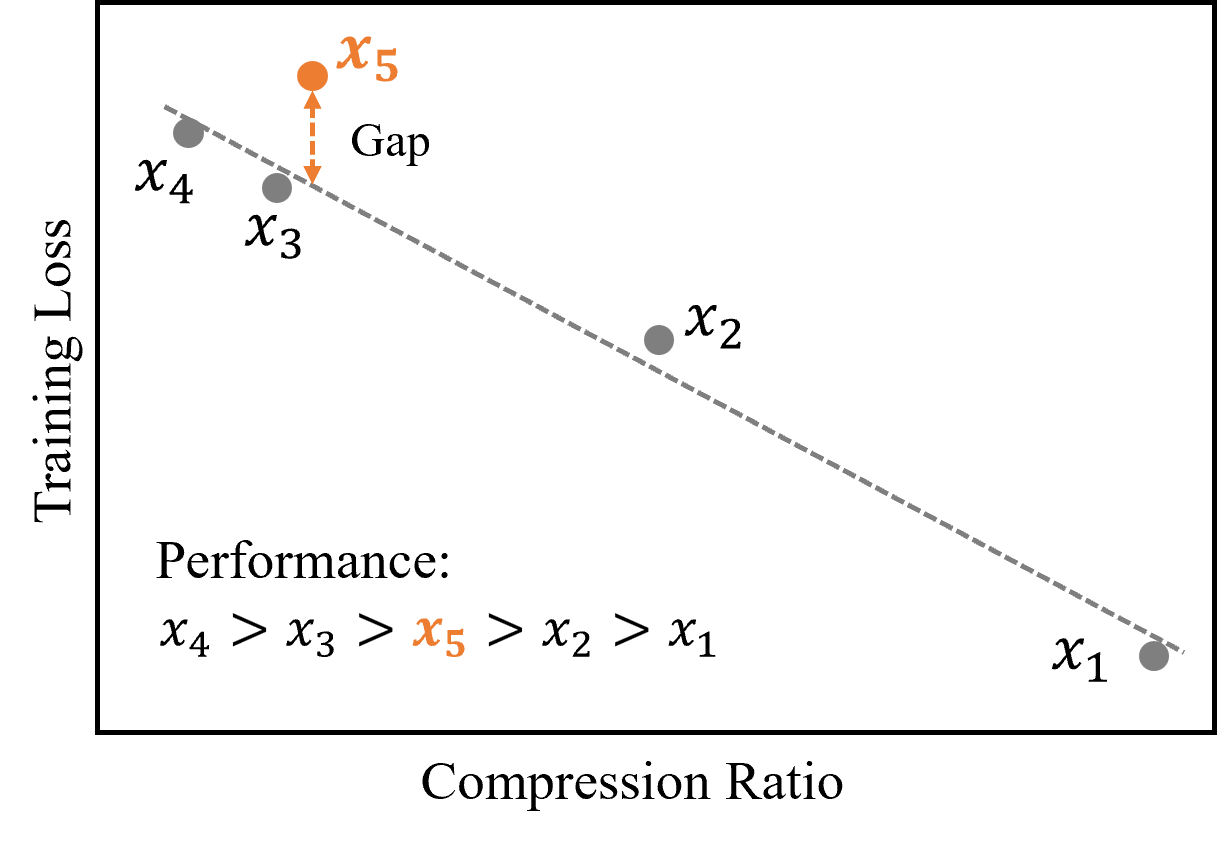}
    \caption{
    Practical application of Entropy law in incremental training data update, where $x_1,x_2,x_3,x_4,x_5$ are five data versions.
    }
    \label{fig: entropy_law_application}
\end{figure}

%% file: Content/6.conclusion.tex
In this paper, we delve deeply into the data selection problem from a data compression perspective.
Inspired by the insight that language modeling is performing information compression, we propose an entropy law delineating the coherent relationship between model performance, data compression ratio, and training loss.
Theoretically guided by the entropy law, we propose a new data selection algorithm, \textbf{ZIP}, to select data with the nearly lowest compression ratio, which is model-free and content-agnostic. rendering it significantly lightweight and versatile.
Experimental results have demonstrated the effectiveness and efficiency of ZIP, based on various LLM backbones, during the SFT and RLHF stages.
Further in-depth analysis provided empirical evidence of Entropy law, which could serve as a criterion for LLM performance prediction on specific data.

%% file: Content/7.appendix.tex
\section{Derivations of joint mutual information of two QA pairs}\label{sec: mutual_information_derivation}

\begin{equation}
\begin{aligned}
I(q_1q_2; a_1a_2) & = H(a_1a_2) - H(a_1a_2|q_1q_2) \\
& = H(a_1a_2) - H(a_1a_2|q_1q_2)\\
& = H(a_1a_2) - H(a_1|q_1q_2) -H(a_2|a_1q_1q_2)\\
& = H(a_1a_2) - H(a_1|q_1) -H(a_2|q_2)\\
& \leq H(a_1) + H(a_2) - H(a_1|q_1) -H(a_2|q_2)\\
& = I(q_1; a_1) + I(q_2; a_2).
\end{aligned}
\end{equation}
The equality is achieved when $a_1$ is independent on $a_2$ (similarly, $q_1$ needs to be independent on $q_2$).

\section{Training Details}\label{sec: training_details}
\textbf{Platform} \space
All experiments were finished on a platform with 64 Intel Xeon Gold 6326 CPU cores @ 2.90GHz, two main-stream high-performance GPUs, and 500GB memories.
The training code is based on a popular open-source framework Axolotl\footnote{\url{https://github.com/OpenAccess-AI-Collective/axolotl}}.

\textbf{Data preprocessing} \space
To format the multi-turn conversation data, we adopt the Vicuna-style template for Mistral-7B and the Llama-3 template for Llama-3-8B.
Samples longer than the maximum input sequence length will be truncated.
Besides, the data will be packed to speed up training for SFT.

\textbf{Hyper-parameters} \space
For ZIP, the selection numbers $K_1$, $K_2$, and $K_3$ are set to 10000, 200, and 100, respectively.
As for SFT, we share these hyper-parameters for all backbones: training batch size is 128, training epochs is 4, input sequence length is 2048, and the warm-up ratio is 0.1.
We adopt different learning rates for each backbone: the learning rate of Mistral-7B is set to 4e-6, and the learning rate of Llama-3-8B is set to 1e-5.
As for RLHF, the learning rate for KTO is set to 1e-6, and the batch size is set to 128.

\section{Token length distribution of more backbones}
The token length distribution of data selected for Llama-3-8B is depicted in Figure \ref{fig: llama3_distribution}, similar to the ones of Mistral-7B.

\input{Figure/llama3_distribution}

\section{Hyper-parameter sensitivity}

\input{Figure/hyper}

ZIP involves three hyper-parameters $K_1$, $K_2$, and $K_3$ for improved efficiency.
We aim to investigate the impact of these hyper-parameters on the model performance, with results depicted in Figure \ref{fig: hyper}.

\textbf{Perceived sample number in global selection $K_1$} \space
$K_1$ decides the number of samples to be updated in the global selection stage.
We set $K_1$ among range [200, 1000, 10000, 20000], and the results are presented in Figure \ref{fig: hyper_k1}.
In the figure, the model performance exhibits an increasing trend when $K_1$ increases.
When a smaller $K_1$ is specified, ZIP is only exposed to a limited set of samples.
This can lead ZIP to degenerate into a variant that consistently selects samples based on individual compression ratios, neglecting the modeling of combinatorial effects. 
Furthermore, the compression ratio associated with the currently selected dataset typically increases with each update, whereas the compression ratios of other samples remain unchanged.
Consequently, a large $K_1$ may result in the compression ratio of the un-updated samples being underestimated, leading to inferior samples' selection.
As a result, a model performance degradation can be found when $K_1$ is set to 20,000.

\textbf{Data pool size of local selection $K_2$} \space
$K_2$ decides the number of samples selected from the previous $K_1$ samples.
We set $K_2$ among range [100, 200, 500, 1000], and the results are presented in Figure \ref{fig: hyper_k2}.
The model performance increases with an increased $K_2$, which aligns with intuition since the algorithm can consider the combinatorial effects of more samples.
But when $K_2$ exceeds a threshold, the model performance reaches a saturated phase, which indicates similar local selection results even with increased local data budget.

\textbf{Data budget of local selection $K_3$} \space
$K_3$ decides the number of samples selected from the previous $K_2$ samples.
We set $K_3$ among range [50, 100, 150, 200], and the results are presented in Figure \ref{fig: hyper_k2}.
The results exhibit a similar trend as the results of $K_1$, yet the underlying causes are inverse.
A large $K_3$ will make ZIP degenerate into a trivial variant that consistently selects samples based on individual compression ratios.
On the other hand, a small $K_3$ will lead to more frequent compression ratio updates, which can also lead to underestimated compression ratios of some inferior samples.

%% file: Figure/llama3_distribution.tex
\begin{figure}[t]
    \centering
    \begin{subfigure}[b]{0.3\textwidth} 
        \includegraphics[width=\textwidth]{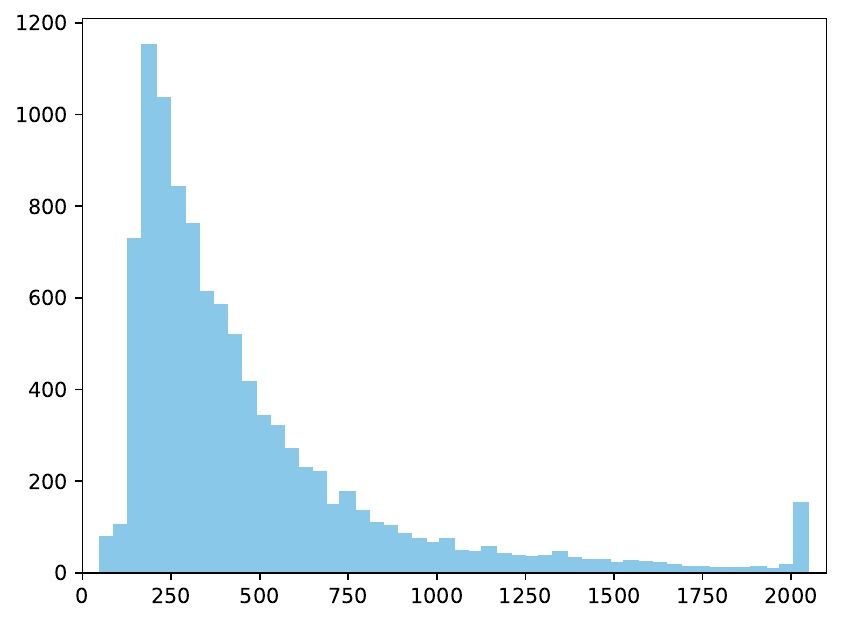}
        \caption{ZIP}
        \label{fig: llama3_distribution_zip}
    \end{subfigure}
    \hfill
    \begin{subfigure}[b]{0.3\textwidth} 
        \includegraphics[width=\textwidth]{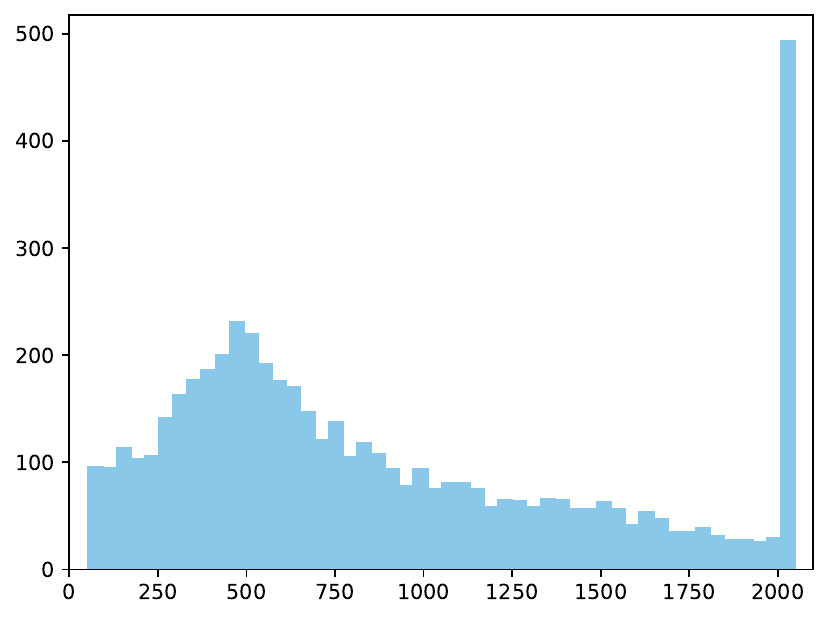}
        \caption{Random}
        \label{fig: llama3_distribution_random}
    \end{subfigure}
    \hfill
    \begin{subfigure}[b]{0.3\textwidth}
        \includegraphics[width=\textwidth]{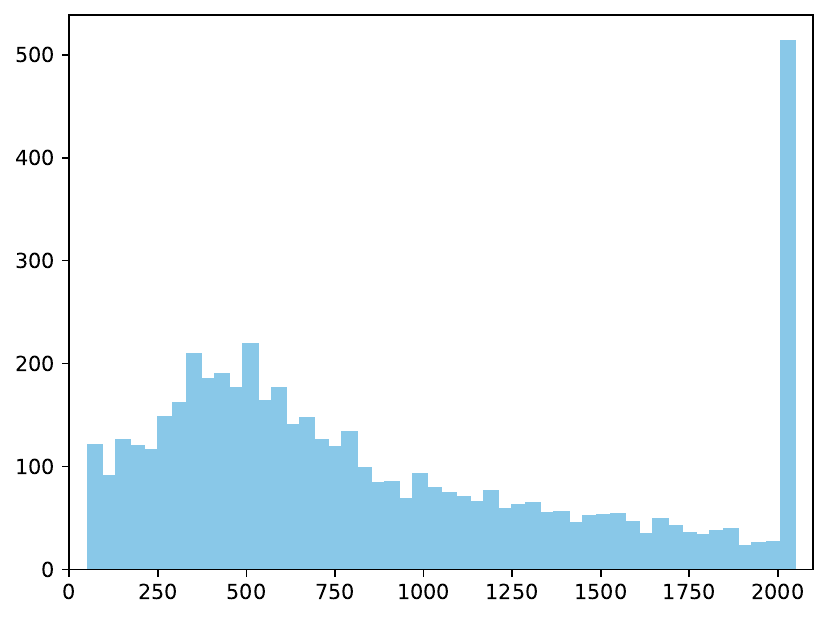}
        \caption{Diversity}
        \label{fig: llama3_distribution_diversity}
    \end{subfigure}

    \vfill 

    \begin{subfigure}[b]{0.3\textwidth} 
        \includegraphics[width=\textwidth]{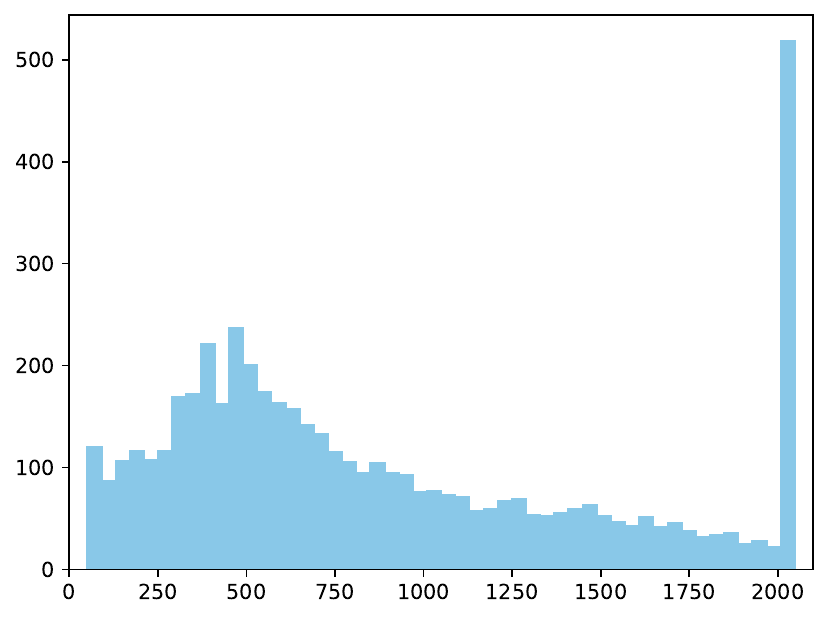}
        \caption{Perplexity}
        \label{fig: llama3_distribution_perplexity}
    \end{subfigure}
    \hfill
    \begin{subfigure}[b]{0.3\textwidth} 
        \includegraphics[width=\textwidth]{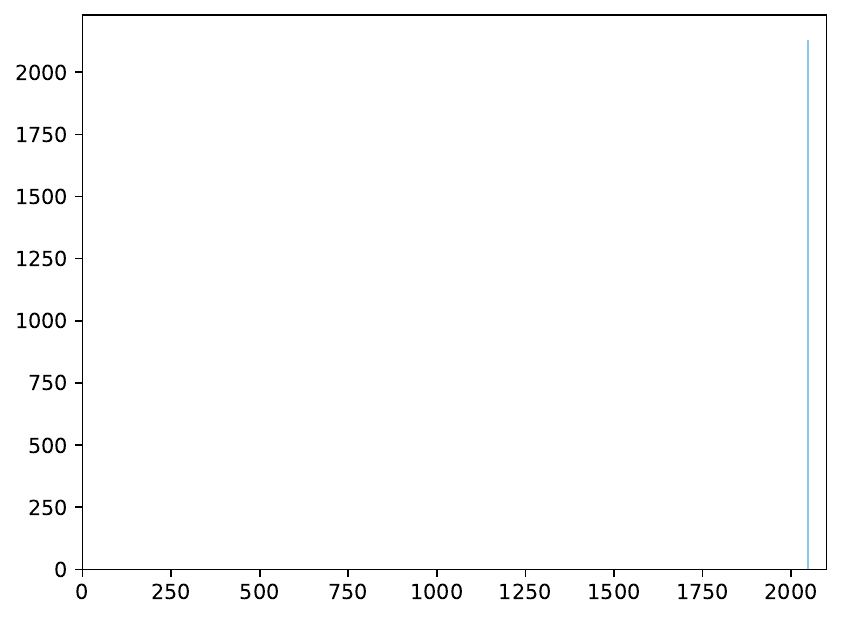}
        \caption{DEITA}
        \label{fig: llama3_distribution_deita}
    \end{subfigure}
    \hfill
    \begin{subfigure}[b]{0.3\textwidth} 
        \includegraphics[width=\textwidth]{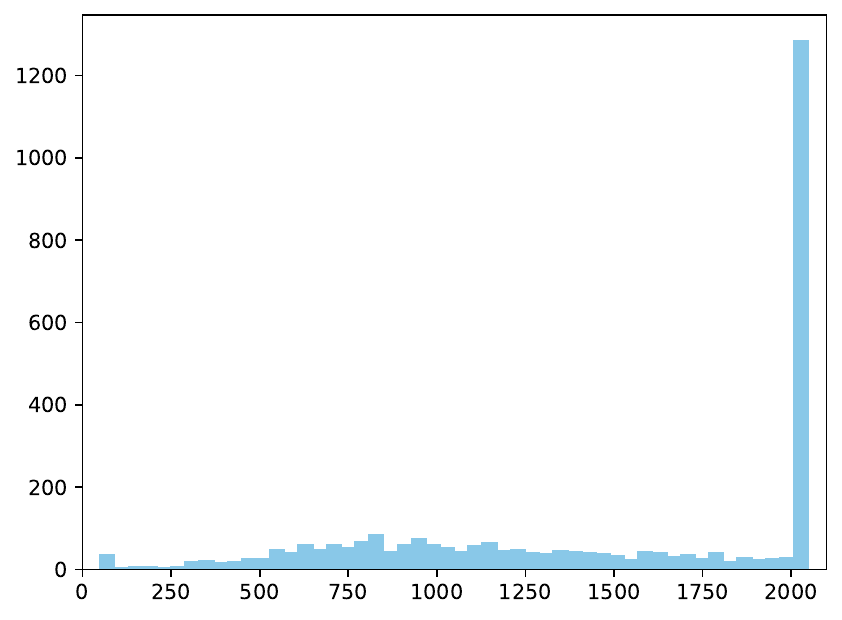}
        \caption{SuperFiltering}
        \label{fig: llama3_distribution_IFD}
    \end{subfigure}
    \caption{The distribution of average token number across datasets selected by different algorithms for Llama-3-8B.}
    \label{fig: llama3_distribution}
\end{figure}

%% file: Figure/hyper.tex
\begin{figure}[t]
    \centering
    \begin{subfigure}[t]{0.32\textwidth} 
        \includegraphics[width=\textwidth]{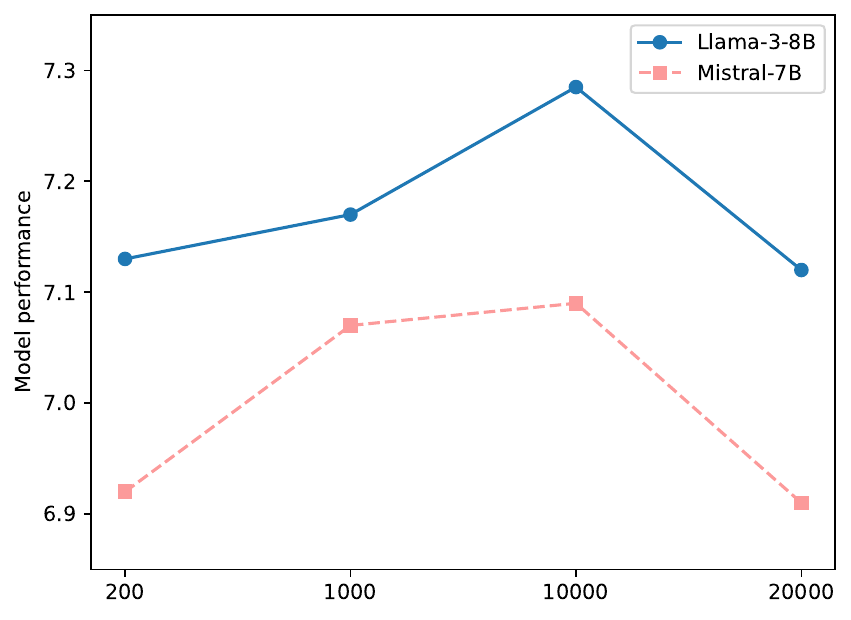}
        \caption{Model performance w.r.t. $K_1$}
        \label{fig: hyper_k1}
    \end{subfigure}
    \hfill
    \begin{subfigure}[t]{0.32\textwidth} 
        \includegraphics[width=\textwidth]{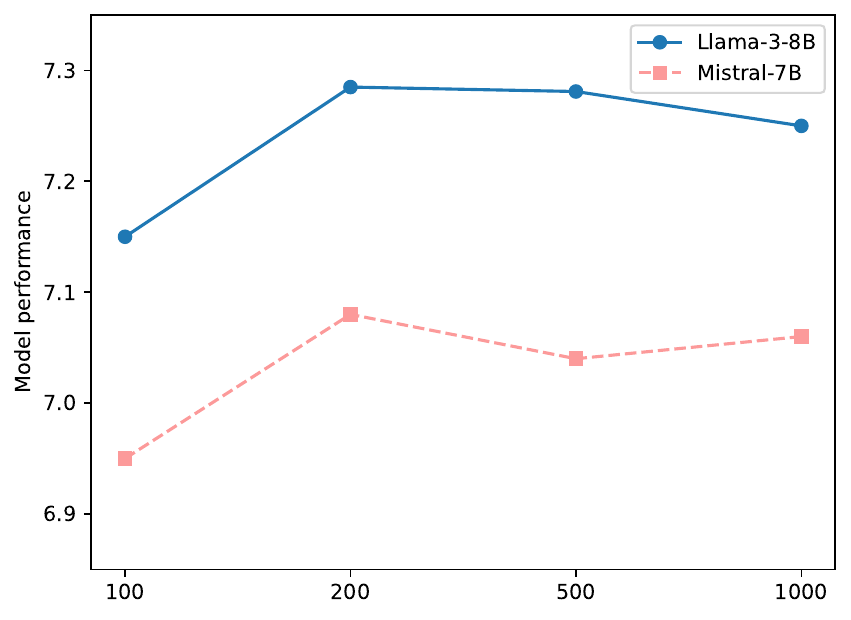}
        \caption{Model performance w.r.t. $K_2$}
        \label{fig: hyper_k2}
    \end{subfigure}
    \hfill
    \begin{subfigure}[t]{0.32\textwidth} 
        \includegraphics[width=\textwidth]{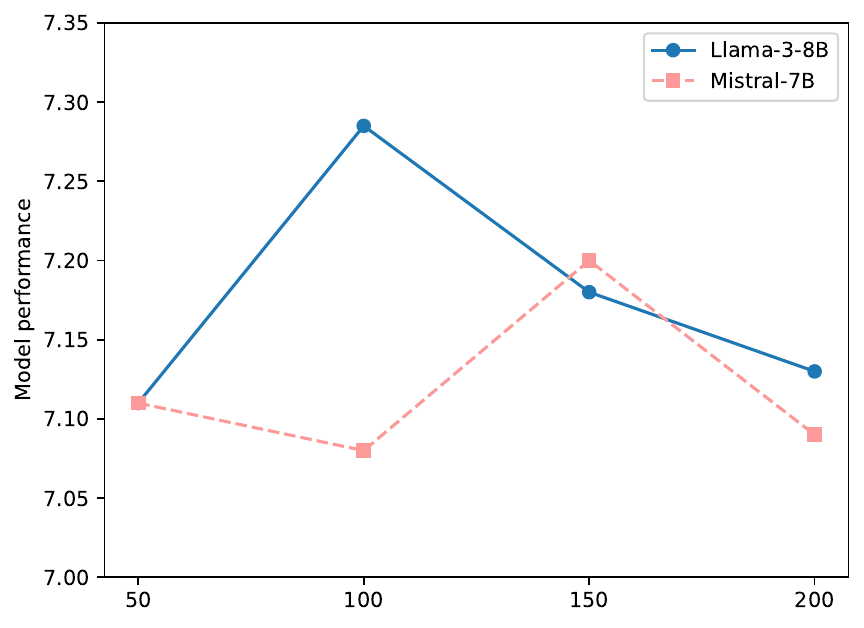}
        \caption{Model performance w.r.t. $K_3$}
        \label{fig: hyper_k3}
    \end{subfigure}
    \caption{Model performance w.r.t. different hyper-parameters.}
    \label{fig: hyper}
\end{figure}